\documentclass[10pt,twocolumn,letterpaper]{article}

\usepackage[pagenumbers]{cvpr} % To force page numbers, e.g. for an arXiv version

\usepackage{algorithm}  
\usepackage{algpseudocode}  
\usepackage{amsmath}
\usepackage{lipsum}
\renewcommand{\algorithmicrequire}{\textbf{Input:}}  % Use Input in the format of Algorithm  
\renewcommand{\algorithmicensure}{\textbf{Output:}} % Use Output in the format of Algorithm  

\usepackage[dvipsnames]{xcolor}

\definecolor{cvprblue}{rgb}{0.21,0.49,0.74}
\usepackage[pagebackref,breaklinks,colorlinks,citecolor=cvprblue]{hyperref}
\usepackage{multirow}
\usepackage{makecell}

\def\paperID{1925} % *** Enter the Paper ID here
\def\confName{CVPR}
\def\confYear{2024}

\title{RS-Corrector: Correcting the Racial Stereotypes in Latent Diffusion Models}
\author{Yue Jiang$^{1,2}$ \quad Yueming Lyu$^{1,2}$ \quad Tianxiang Ma$^{1,2}$\quad Bo Peng$^{1}$\quad Jing Dong$^{1*}$\\
$^{1}$Institute of Automation, CAS\\
$^{2}$School of Artificial Intelligence, UCAS\\
}
\newcommand{\themodel}{RS-Corrector\xspace}

\newcommand{\annotator}[2]{\csdef{#1}##1{{\color{#2} [\textbf{\MakeUppercase #1}: ##1]}}}
\annotator{jy}{red}

\begin{document}
\twocolumn[{%
% \maketitle
\maketitle
\begin{figure}[H]
\hsize=\textwidth % cvpr 需要
\centering
\includegraphics[width=1\textwidth]{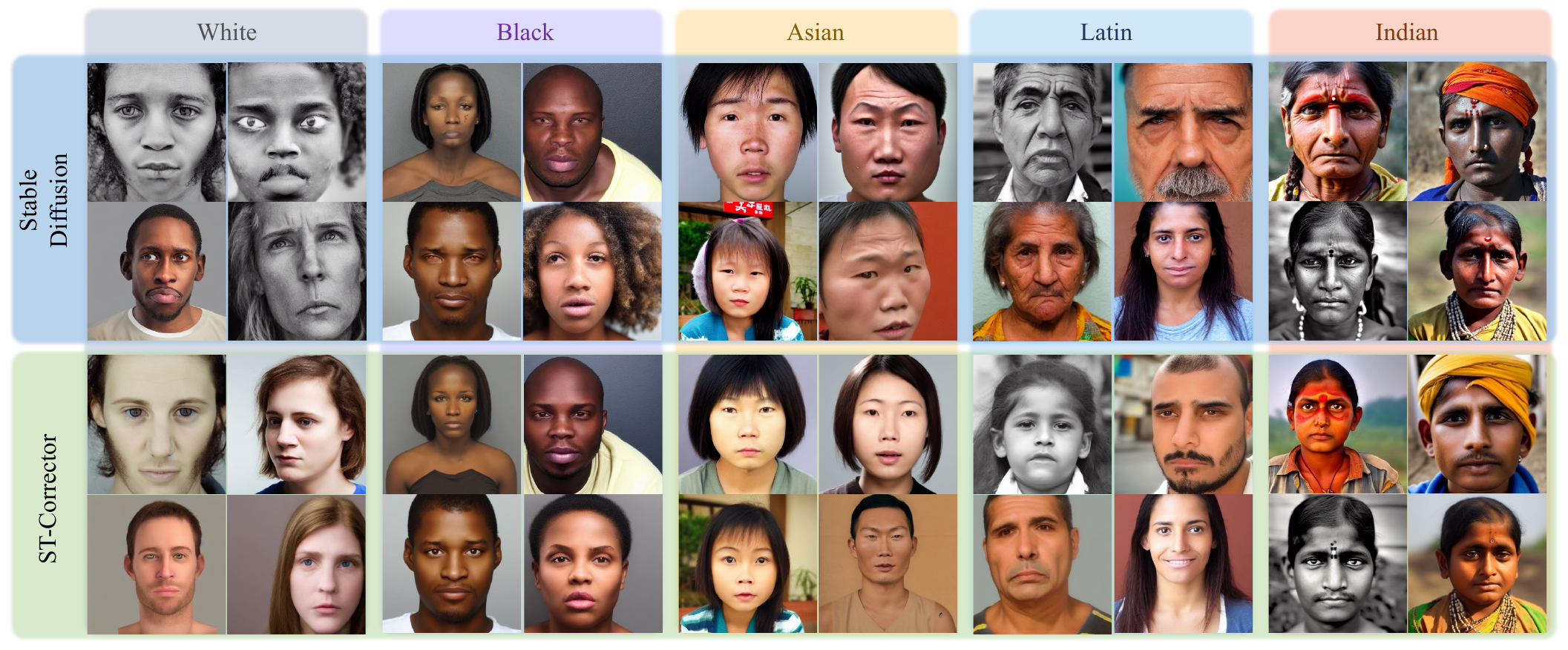}
\caption{Given a pre-trained latent diffusion model (LDM) (e.g., Stable Diffusion~\cite{rombach2022high}), which exhibits apparent stereotypes for different racial groups, our RS-Corrector establishes an anti-stereotypical preference, which guides the adjustment of the latent code during the inference stage. Consequently, the stereotypes are corrected and the generated results exhibit significant improvements with respect to facial appearance.}
% obtained anti-stereotypical preferences.}
%in the latent space and update the latent code for refined generated results without fine-tuning or re-training the original model.}
\label{fig:teaser}
\end{figure}
}]
% \renewcommand{\thefootnote}{}
% \footnote{$^{*}$ Corresponding author.}
% \documentclass{article}
% \usepackage{lipsum}
 
\newcommand\blfootnote[1]{%
  \begingroup
  \renewcommand\thefootnote{}\footnote{#1}%
  \addtocounter{footnote}{-1}%
  \endgroup
}
 
% \begin{document}
 
% Some text\blfootnote{A footnote witout marker} and some more text\footnote{A standard footnote}
 
% \end{document}
\blfootnote{$^{*}$ Corresponding author.}
% \footnote{\noindent
%         \textbf{收稿日期}：text\\ 
%         \textbf{基金项目}：text\\ 
%         \textbf{作者简介}：text\\
% }

\begin{abstract}
Recent text-conditioned image generation models have demonstrated an exceptional capacity to produce diverse and creative imagery with high visual quality.
However, when pre-trained on billion-sized datasets randomly collected from the Internet, where potential biased human preferences exist, these models tend to produce images with common and recurring stereotypes, particularly for certain racial groups. In this paper, we conduct an initial analysis of the publicly available Stable Diffusion model and its derivatives, highlighting the presence of racial stereotypes. These models often generate distorted or biased images for certain racial groups, emphasizing stereotypical characteristics. To address these issues, we propose a framework called ``\textbf{\themodel}'', designed to establish an anti-stereotypical preference in the latent space and update the latent code for refined generated results. The correction process occurs during the inference stage without requiring fine-tuning of the original model. Extensive empirical evaluations demonstrate that the introduced \themodel effectively corrects the racial stereotypes of the well-trained Stable Diffusion model while leaving the original model unchanged. 
\end{abstract}

\section{Introduction}
\label{sec:intro}
% 越来越多的研究者发现SD生成的结果存在各种不合适的情况，包括暴力，色情，种族歧视和固有偏见，
In recent years, text-conditioned generative models~\cite{saharia2022photorealistic,rombach2022high,ramesh2022hierarchical,zhang2023adding,hu2021lora,kawar2023imagic} have demonstrated impressive capabilities, showing significant promise for various downstream applications such as style transfer~\cite{zhang2023prospect,tumanyan2023plug,li2023blip,shi2023dragdiffusion}, concept learning~\cite{ruiz2023dreambooth,ruiz2023hyperdreambooth,kumari2023multi}, image-restoration~\cite{fei2023generative,wang2023dr2} and so on.

Unfortunately, these pretrained generative models have been demonstrated to present degraded and biased behaviors, such as generating inappropriate content, including insults, threats, and explicit material, or exhibiting stereotypes in terms of race, gender, and occupation~\cite{schramowski2023safe,howard2023probing,luccioni2023stable,friedrich2023fair,naik2023social}. 
This can be attributed to their training on large-scale unfiltered datasets, like LAION-5B~\cite{schuhmann2022laion}, which may contain biases or harmful content~\cite{birhane2021multimodal,friedrich2023fair}. Consequently, the resulting models have the potential to reproduce unexpected behaviors, leading to harm in downstream tasks.

\begin{figure}[t]
    \centering
    \includegraphics[width=\linewidth]{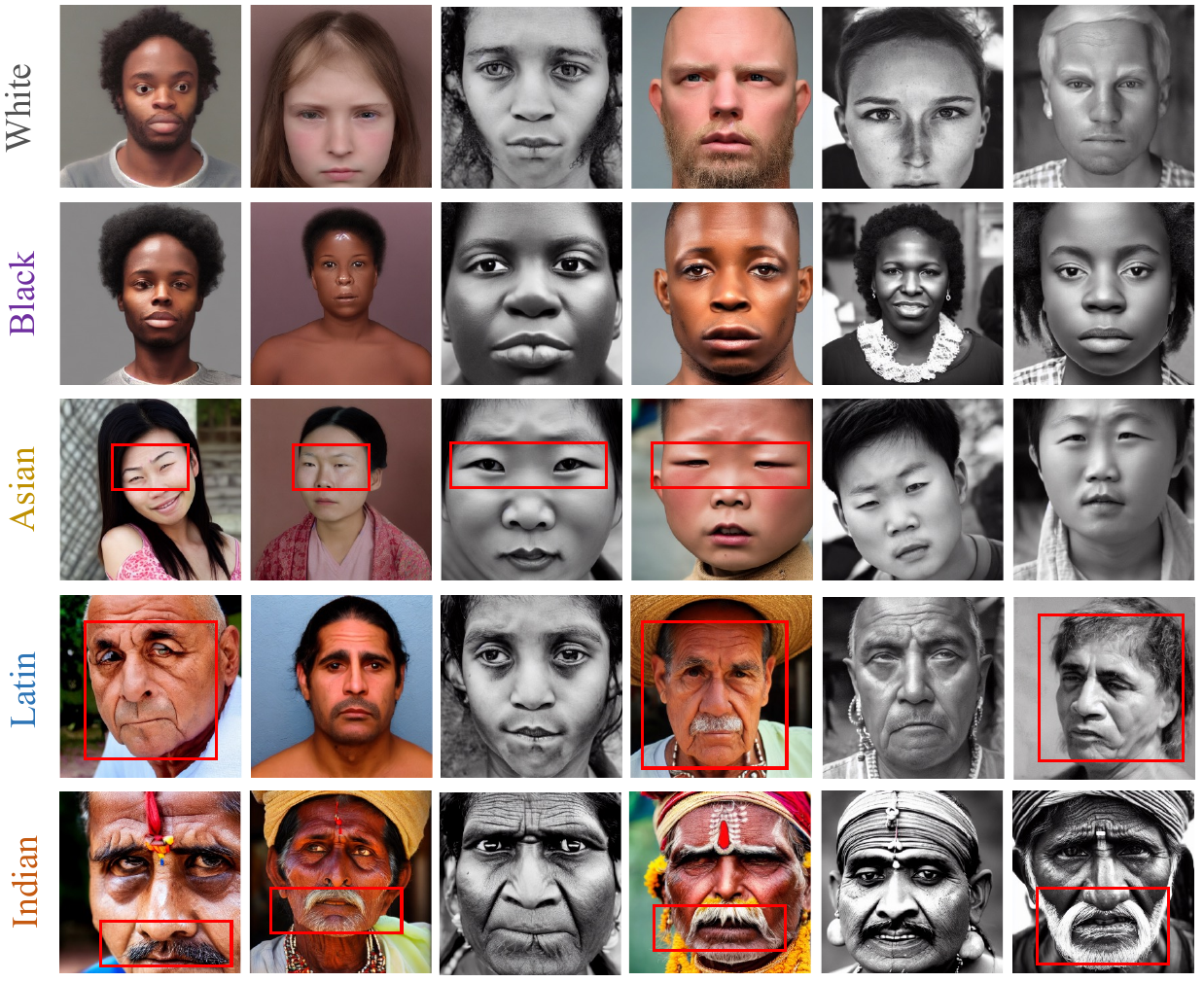}
    \caption{Illustration of racial stereotypes in Stable Diffusion (SD)~\cite{rombach2022high}. The generated results of the Asian group suffer from defamatory in terms of facial aesthetics, while the results of Latin and Indian groups exhibit signs of apparent aging when compared to others, which demonstrates multiple racial stereotypes in the model.}
    \label{fig:sd_gen}
\end{figure}

However, addressing the intrinsic stereotypes of the generative models is crucial but currently underexplored.
In this paper, we explore the state-of-the-art text-conditional image generation models, such as Stable Diffusion~\cite{rombach2022high}, and observe that it easily exhibits racial stereotypes, $\emph{i.e.}$, the results generated for different racial groups tend to produce specific stereotypical characteristics. As shown in Fig.~\ref{fig:sd_gen}, when the prompt is ``a face of an Asian'', the model produces images with common stereotypes such as small eyes, high cheekbones, and long middle face, which presents an unflattering portrayal of this racial group. Likewise, the prompt given ``a face of an Indian'' tends to result in images with noticeable wrinkles, reinforcing another racial stereotype.

To solve this problem, 
we propose an effective method named ``RS-Corrector'' to correct the racial stereotypes of latent diffusion models. Our main objective is to acquire an anti-stereotypical latent prior in the latent space and utilize it to adjust the latent code for a fairer generation. To achieve this, we initially learn to classify anti-stereotypical and stereotypical representations using a collected fair dataset and the generated stereotypical images. Given the recent scarcity of racially fair data, we construct a high-quality dataset on various racial groups avoiding the introduction of additional biases including gender and age. With the proposed anti-stereotypical data, we optimize and obtain the latent prior via a contrastive manner.
Subsequently, the acquired latent prior is applied during the inference stage to guide the update of the latent code towards the preferred anti-stereotypical direction. With our approach, the generated results are rectified, displaying fairer outcomes. As depicted in Fig.~\ref{fig:teaser}, the facial appearance of different racial groups is largely improved.

To summarize, our contributions are as follows:

\begin{itemize}
\item 
We reveal the racial stereotypes in the state-of-the-art Stable Diffusion model and construct a fair dataset in terms of different racial groups eliminating biases related to other attributes, notably gender and age.
\item
We introduce an effective framework called ``RS-Corrector'' to correct the racial stereotypes ingrained in the original model, which focuses on obtaining an anti-stereotypical prior in the latent space and updating the latent code towards refined generated results. Our method necessitates no modification to the original model and requires short inference time.
\item
Extensive qualitative and quantitative experiments demonstrate that our method is superior at correcting racial stereotypes. The generated results of RS-Corrector are plausible in terms of facial aesthetics and attribute fairness.
% , which makes substantial contributions to societal harmony.
\end{itemize}

\section{Related Work}
\label{sec:related_work}

\textbf{Text-guided Image Generation.} Text-guided image generation aims at translating textual description into images, which early centered on Variational AutoEncoders (VAEs)~\cite{kingma2013auto,esser2021taming} and Generative Adversarial Networks (GANs)~\cite{xia2021tedigan,gal2022stylegan,patashnik2021styleclip,lyu2023deltaedit}.
% 换一下语序
Recently, diffusion models have achieved more impressive and high-quality results, with higher super-resolution and more stable training process~\cite{kim2022diffusionclip,avrahami2022blended,hertz2022prompt,ramesh2021zero,gu2023matryoshka,xiao2023fastcomposer,lee2023aligning,bar2023multidiffusion}. However, training a diffusion model can be resource-intensive, as it involves predicting noise images in a series of timesteps in pixel space. To address this issue, Rombach~\emph{et al.} propose Stable Diffusion Model~\cite{rombach2022high} which is trained on the latent space of reduced dimensionality, effectively mitigating the problem of high training costs while ensuring the generation performance uncompromised. SD has also become one of the most favored pipelines in the research community, exhibiting its great potential for multiple applications, such as image editing~\cite{hertz2022prompt,preechakul2022diffusion,mokady2023null,brooks2023instructpix2pix,lyu2023deltaspace,meng2021sdedit}, transfer learning~\cite{parmar2023zero,zhang2023inversion,shi2023dragdiffusion} and concept learning~\cite{ruiz2023dreambooth,kumari2023multi,yuan2023inserting}. These models are mostly fine-tuned on Stable Diffusion and can achieve more specific tasks.  DreamBooth~\cite{ruiz2023dreambooth} serves as a classical fine-tuning work focusing on learning a new concept within several images. It fine-tunes the whole model
% , while Custom Diffusion fine-tunes a subset of the parameters in the cross-attention layers. Both of them can 
to generate a sequence of high-quality images pertaining to a new concept. Textual Inversion~\cite{gal2022image} is proposed to obtain unique and varied concepts into a single word embedding,
% a kind of prompt engineering approach to learn a new style or object, it obtains a new embedding of the target 
without any modification of the main part of diffusion models. However, these text-conditioned diffusion models do not address the issue of model inherited stereotypes. 
% are discovered to present severe unfair and inappropriate generation results.

\textbf{Potential Threats of Diffusion Models.} \quad Unfortunately, as many diffusion models are trained on large-scale unfiltered datasets, LAION-5B for example, containing various problematic content including sensitive topic such as rape, porngography, malign stereotypes, racism and so on~\cite{birhane2021multimodal}, these models tend to present degraded behavior
% experience rapid advancements, an increasing number of researchers have identified potential challenges associated with the generated results
across various aspects. Schramowski~\emph{et al.}~\cite{schramowski2023safe} reveal that SD~\cite{rombach2022high} can generate images of pornography, violence, panicky and so on. Luccioni~\emph{et al.} and Friedrich~\emph{et al.}~\cite{luccioni2023stable,friedrich2023fair} discover that diffusion models exhibit unfairness with regard to specific occupation in gender. Moreover, Naik~emph{et al.}~\cite{naik2023social} systematically analyze the fairness issues within DALLE-v2~\cite{ramesh2022hierarchical} and SD, considering four social bias dimensions: gender, race, age and geographical location by prompts of occupations, personality traits, and everyday situations. Their findings reveal a propensity in both models to generate images of a certain group, which has the potential to reinforce stereotypes or perpetuate social bias. In this paper, we further analyze the performance of SD across different racial groups
% as well as several downstream tasks 
to unveil the potential challenges and concerns. 

\textbf{Optimizing LDMs during the Inference Stage.} 
Optimizing LDMs during the inference stage usually involves updating the latent code, which does not fine-tune the original model~\cite{samuel2023all,samuel2023norm,chefer2023attend,li2023divide}. SeedSelect~\cite{samuel2023all} focuses on optimizing the latent code to an optimal initial noise to achieve rare concept generation. However, it requires long generation times, yielding computational limitations. 
% However, its results are confined to the performance of original space, which is further improved by
NAO~\cite{samuel2023norm} finds a $\chi$ distribution over the norms, which helps improve the generation ability and efficiency. Attend-and-Excite~\cite{chefer2023attend} optimizes the latent code for more precise alignment between textual conditions and generated images. It focuses on the probability distribution of text tokens for each image patch, which is obtained from the attention maps. However, these methods do not specifically concentrate on rectifying the inherent stereotypes within the model.

\section{Method}
\subsection{Preliminaries on Stable Diffusion}
\quad Our method is based on Stable Diffusion~\cite{rombach2022high}, which iteratively performs denoising in the latent space of an autoencoder, namely $\mathcal{E}(\cdot)$ and $\mathcal{D}(\cdot)$. The encoder $\mathcal{E}(\cdot)$ first encodes an image $x$ into a latent representation $z={\mathcal{E}}(x)$ and the decoder $\mathcal{D}(\cdot)$ reconstructs image $\Tilde{x}$ from $z$, giving $\Tilde{x}=\mathcal{D}(z)=\mathcal{D}(\mathcal{E}(x))$. 

In the training stage, a latent code $z_t$ in the latent space is denoised by a denoising diffusion probabilistic model (DDPM)~\cite{sohl2015deep,ho2020denoising} at each timestep t. Specifically, for text-to-image generation, an additional text embedding $c(y)$ is added to the model by the CLIP text encoder~\cite{radford2021learning} from the conditioned prompt $y$. The DDPM model $\epsilon_{\theta}$ is trained by minimizing the following loss,
% Deep unsupervised learning using nonequilibrium thermody
% Denoising diffusion probabilistic models
\begin{equation}
\mathcal{L}=\mathbb{E}_{z\sim\mathcal{E}(x),y,\epsilon\sim\mathcal{N}(0,I),t}\left[\left \|\epsilon-\epsilon_\theta(z_t,t,c(y))\right\|_2^2\right],
\end{equation}
% Learning transferable visual models from natural language supervision
where $t$ is the timestep, as denoising steps are operated step by step.

In the inference stage, a latent noise $z_T$ is sampled from the standard Gaussian distribution $\mathcal{N}(0,I)$ and iteratively denoised by the pretrained model $\epsilon_{\theta}$. The latent code $z_0$ is obtained after the denosing process for $T$ timesteps and is subsequently fed into the image decoder $\mathcal{D}(\cdot)$ to produce the generated image $I=\mathcal{D}(z_0)$. 
% 加不同阶段噪声加入的影响

\begin{figure}[t]
    \centering
    \includegraphics[width=\linewidth]{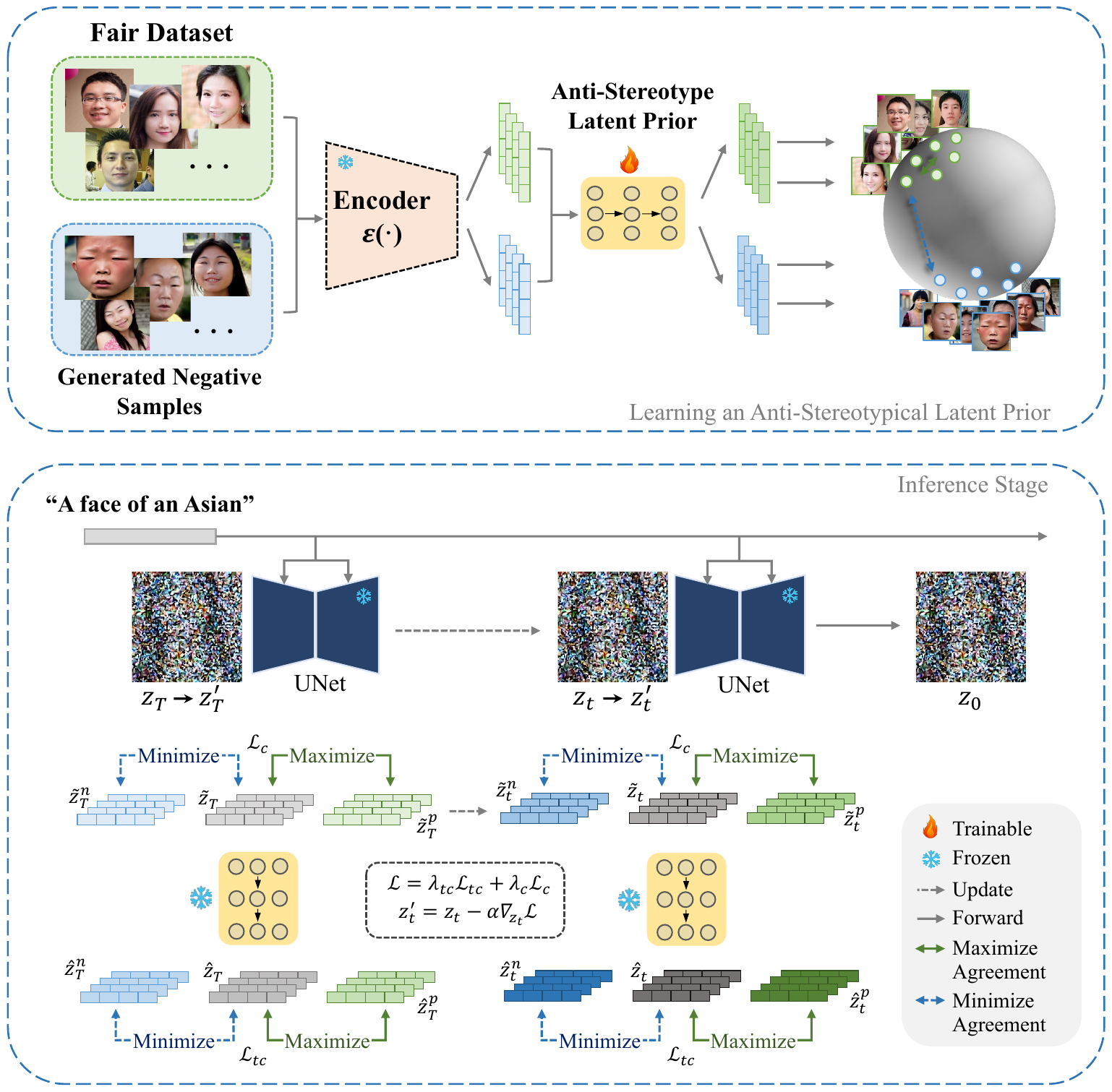}
    \caption{Overview of RS-Corrector. Given a fair dataset and generated negative samples, we establish an anti-stereotype latent prior and employ it to guide the adjustment of the latent code, cooperating with direct anti-stereotypical guidance. Notably, the latent code undergoes iterative updates over a specific number of timesteps during the inference stage.}
    \label{fig:framework}
\end{figure} 
\subsection{\themodel}
In this section, we introduce the proposed \textbf{\themodel}, which aims to correct the intrinsic stereotype without fine-tuning the original model. The key idea is to obtain an anti-stereotypical preference in the latent space and update the latent code towards a fairer generation. The whole process is conducted in the inference stage, obviating the necessity for any modification to the original model. We leverage the Stable Diffusion model as our generative backbone. Note that our method is not confined to a certain model, it functions as a comprehensive rectification approach applicable to a range of LDMs.

To obtain the latent preference, we first construct a latent prior to capture the most distinguishable representations between anti-stereotypical and stereotypical samples generated by Stable Diffusion. This latent prior is then employed in the inference stage to guide the adjustment of the latent code towards the preferred anti-stereotypical direction. To further strengthen this process, we introduce a direct anti-stereotypical guidance to coherently update the latent code. Both of these constraints work in tandem to steer the latent code towards representations that closely resemble the anti-stereotypical representation, beneficial for a fairer generation.

\textbf{Learning an Anti-stereotypical Latent Prior.} 
To correct the stereotype during the inference stage, we identify the most distinctive representations of anti-stereotypical and stereotypical images in the latent space as a latent prior.

Note that the anti-stereotypical images should be fair for public election. Since such datasets are lacked recently, we build a dataset encompassing various racial groups with equally distributed attributes in terms of gender and age to eliminate their influence. The data is initially filtered from LAION-5B~\cite{schuhmann2022laion} by a face detection model libface\footnote{\url{ https://github.com/ShiqiYu/libfacedetection}}  and labeled by CLIP~\cite{radford2021learning}, and further validated by human workers.

To obtain the latent prior, we consider utilizing a pertrained classifier $\mathcal{F}(\cdot)$ in the latent space with $\mathcal{F}(z_t)=c$, where $c$ is the condition for anti-stereotype. Then the conditioned generation result $\epsilon_\theta(z_t|c) $ can be obtained as $\epsilon_\theta(z_t|c) \varpropto p(\mathcal{F}|z_t)\epsilon_\theta(z_t)$, where $p(\mathcal{F}|z_t)=p(\mathcal{F}(z_t)=c|z_t)$~\cite{sinha2021d2c}. $\epsilon_\theta$ is the noise predictor of the latent diffusion model and $z_t$ is the latent code in the inference stage. $\mathcal{F}(z_t)=c$ is regard as the latent prior. Latent code $z_t$ can be updated to $\hat{z_t}$ as follow:
\begin{equation}
    \hat{z_t} \sim \mathcal{F}(c|z_t) \cdot \epsilon_\theta(z_t).
\end{equation}

For given anti-stereotypical images $I_p$, which are considered to be ``positive'', stereotypical images $I_n$ generated by SD are regarded as ``negative''. The two types of images are sent to the image encoder $\mathcal{E}$ of SD for latent representation $z^p_0$=$\mathcal{E}(I_p)$ and $z^n_0$=$\mathcal{E}(I_n)$ to guarantee spatial consistency in the latent space. Our goal is to conduct a latent prior $F$ for $z^p_0$ and $z^n_0$. In order to enhance the extraction of discerning features and facilitate the computation of contrastive similarity, the latent representations $z\in R^{B\times C\times H\times W}$ of four dimensions is first flattened to $R^{B\times(C\times H\times W)}$ of two dimenstions~\cite{oord2018representation}. Additionally, as feature normalization is crucial for feature combination and classification~\cite{wang2020understanding}, the flattened representations are then normed to a unit vector $\Tilde{z}$ as following:
\begin{equation}
\begin{aligned}
\label{eq:z_0}
% z^p_0 & =\mathcal{E}(I_p),\\
% z^n_0 & =\mathcal{E}(I_n),\\
&\Tilde{z}^p_0 =Norm(flatten(\mathcal{E}(I_p))),\\
&\Tilde{z}^n_0 =Norm(flatten(\mathcal{E}(I_n))).
\end{aligned}
\end{equation}

\begin{figure*}[t]
    \centering
    \includegraphics[width=\linewidth]{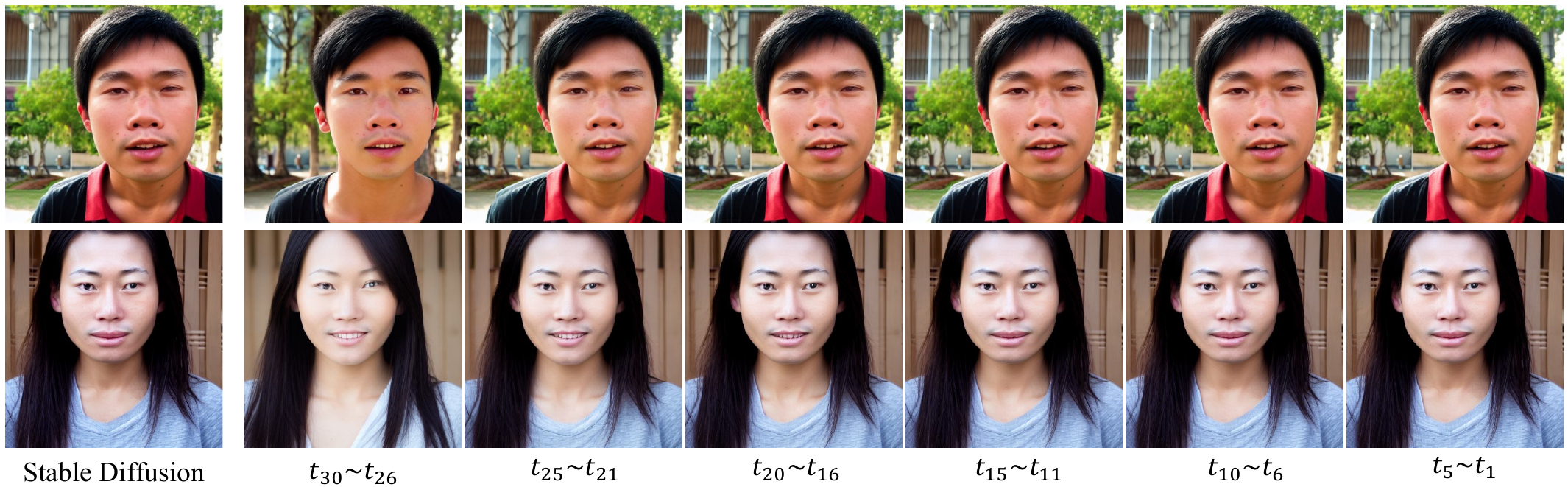}
    \caption{Illustration of updating the latent code at different timesteps out of 30 total timesteps. The Adjustment in the later timesteps yields more effective content modifications.}
    \label{fig:optimazation}
\end{figure*}

The latent prior is conducted by adding a projector to the image encoder $\mathcal{E}$ as $F(\Tilde{z})$, which only consists of a single fully connected layer. New unit vectors $\hat{z}_p$, $\hat{z}_n$ is obtained as $\hat{z}_p=F(\Tilde{z}^p_0),\hat{z}_n=F(\Tilde{z}^p_0)$. The projector is trained by maximizing the agreement between positive pairs and minimizing the agreement between negative pairs~\cite{oord2018representation} as follows:
\begin{equation}
\begin{aligned}
\label{eq_fc}
&\mathcal{L}_{F}=-log\frac{\hat{s}_p^{i,j}}{\hat{s}_p^{i,j}+\sum_m \hat{s}_n^{i,m}},\\
&\hat{s}_p^{i,j}=exp(sim(\hat{z}_p^i,\hat{z}_p^j)/ \tau ),\\
&\hat{s}_n^{i,m}=exp(sim(\hat{z}_p^i,\hat{z}_n^m)/ \tau),
\end{aligned}
\end{equation}
where $\tau$ is a temperature parameter and $sim(\cdot, \cdot)$ is the similarity function which is the dot product in this work.

The trained projector presents as the anti-stereotypical latent prior to guide the adjustment of the latent code for fairer generation.

\textbf{Obtaining a Direct Anti-stereotypical Guidance.}\quad To further guide the adjustment of the latent code, we employ a direct contrastive constraint during the inference stage. For a latent code $z_i$, we randomly select a positive sample $I_p$ and a negative sample $I_n$ and obtain their latent representation $z^p_0=\mathcal{E}(I_p)$, $z^n_0=\mathcal{E}(I_n)$. The noisy version $z^p_t, z^n_t$ of $z^p_0, z^n_0$ at timestep $t$ can be directly sampled as follows:
\begin{equation}
    \begin{aligned}
    \label{eq:z_t}
        &z^{p}_t=\sqrt{\alpha_t}\cdot z^{p}_0+\sqrt{1-\alpha_t}\cdot z^{p}_0\epsilon,\\
        &z^{n}_t=\sqrt{\alpha_t}\cdot z^{n}_0+\sqrt{1-\alpha_t}\cdot z^{n}_0\epsilon,
    \end{aligned}
\end{equation}
where $\alpha_t$ is a fixed scale factor, and $\epsilon$ is a Gaussian noise. After flattened and normalized, the similarity between $\Tilde{z}_t$ and $\Tilde{z}_t^p$ in a unit hypersphere is obtained by a traditional contrastive loss as follows:
\begin{equation}
\begin{aligned}
\label{eq:l_c}
&\mathcal{L}_{c}=-log\frac{\Tilde{s}_p^{t,p}}{\Tilde{s}_p^{t,p}+\Tilde{s}_n^{t,n}},\\
&\Tilde{s}_p^{t,p}=exp(sim(\Tilde{z}_t,\Tilde{z}_t^p)/ \tau^{'}),\\
&\Tilde{s}_n^{t,n}=exp(sim(\Tilde{z}_t,\Tilde{z}_t^n)/ \tau^{'}),
\end{aligned}
\end{equation}
where $\tau^{'}$ is a different temperature parameter to $\tau$ in Eq.~\ref{eq_fc}.

The purpose of both losses is to instruct the adjustment of the latent code towards the direction of similarity with positive samples in the latent space.

In the inference stage, the pretrained prior $F$ is utilized to obtain prior constraint as follow:
\begin{equation}
    \begin{aligned}
        \label{eq:l_tc}
&\mathcal{L}_{tc}=-log\frac{\hat{s}_p^{t,p}}{\hat{s}_p^{t,p}+\hat{s}_n^{t,n}},\\
    \end{aligned}
\end{equation}
where $\hat{s}_p^{t,p}$ is calculated similar to $\Tilde{s}_p^{t,p}$ in Eq.~\ref{eq:l_c}, but with the use of projected variables $\hat{z}_t=F(\Tilde{z}^t)$, $\hat{z}_t^p=F(\Tilde{z}^p_t)$. $\hat{s}_n^{t,n}$ is calculate in like manner.
The overall objective consists of the above two losses:
\begin{equation}
\begin{aligned}
\label{eq:l_all}
&\mathcal{L}=\lambda_{tc}\mathcal{L}_{tc}+\lambda_c\mathcal{L}_c.\\
\end{aligned}
\end{equation}
Finally, the latent code is updated as follow during the inference stage:
\begin{equation}
    \begin{aligned}
        \label{eq:update}
   &z_t^{'}=\epsilon_\theta (z_t,\mathcal{P},t)-\eta\nabla_{z_t}\mathcal{L},
    \end{aligned}
\end{equation}
where $\mathcal{P}$ is the conditional textual prompt, $t$ is the denoising timestep, and $\eta$ is the optimzation step size. The update of the latent code in each denoising step is detailed in Algorithm~\ref{alg:alg1}.

Moreover, we observe that the adjustment process at different timesteps yields different modifications to the results. Updating the latent code in the later steps results in more pronounced content changes. While in the early steps, the impacts are minimal. This is because the core content and structural attributes are firmly established in later steps during inference. Fig.~\ref{fig:optimazation} illustrates how the adjustment of different timesteps influences the final results. Consequently, to achieve better results more fleetly and improve the model efficiency, we update the latent code over the last few timesteps to eliminate stereotypes as they are being constructed, which exerts more notable influence on the results in a short time.

\begin{algorithm}[htb]
    \renewcommand{\algorithmicrequire}{\textbf{Input:}}
    \renewcommand{\algorithmicensure}{\textbf{Output:}}
  \caption{A Single Denoising Step using \themodel}
  \label{alg:alg1}
  \begin{algorithmic}[1]
    \Require 
    % to adjust the update scale of the latent code and control the tradeoff between different losses
A text prompt $\mathcal{P}$, positive samples from fair dataset $\left\{I_1,...,I_P\right\}$, negative samples generated by SD $\left\{I_1,...,I_N\right\}$, a timestep $t$, a latent noise $z_t$, a set of rectification timesteps $\left\{t_i,...,t_k\right\}$, hyperparameters $\lambda_{tc},\lambda_{c}$, a pretrained SD $\epsilon_\theta$ with image encoder $\mathcal{E}( \cdot )$, a pretrained latent prior $F( \cdot )$.
    \Ensure
    A latent noise $z_{t-1}$ for the next timestep.
   % \State

   \If {${t\in\left\{t_i,...,t_k\right\}}$}

\State
obtain $z_0^{p},z_0^{n}$ encoded from positive and negative samples by $\mathcal{E}(\cdot)$ \hfill $\triangleright Eq.~\ref{eq:z_0}$ 

\State
obtain $z_t^{p},z_t^{n}$ sampled by $\epsilon_\theta$ from $z_0^{p},z_0^{n}$ \hfill $\triangleright Eq.~\ref{eq:z_t}$

\State
obtain $\Tilde{z}_t^p$, $\Tilde{z}_t^n$ flattened and normalized from $z_t^p$,$z_t^n$ 

\State
compute $\mathcal{L}_c$ \hfill $\triangleright Eq.~\ref{eq:l_c}$

\State
$\hat{z}_t^{p},\hat{z}_t^n \leftarrow F(\Tilde{z}_t^p),F(\Tilde{z}_t^n)$ \hfill $\triangleright$ projected by prior $F(\cdot)$ 

\State
compute $\mathcal{L}_{tc}$ \hfill $\triangleright Eq.~\ref{eq:l_tc}$

\State
$\mathcal{L}\leftarrow\lambda_{tc}\mathcal{L}_{tc}+\lambda_{c}\mathcal{L}_c$ \hfill $\triangleright Eq.~\ref{eq:l_all}$

\State
   $z^{'}_{t}{\leftarrow}\epsilon_\theta (z_t,\mathcal{P},t)-\eta\nabla_{z_t}\mathcal{L}$ \hfill $\triangleright Eq.~\ref{eq:update}$
   \State
   {$z_{t-1} \leftarrow \epsilon_\theta (z^{'}_t,\mathcal{P},t)$}
   \Else 
   \State
   {$z_{t-1} \leftarrow \epsilon_\theta (z_t,\mathcal{P},t)$}
   \EndIf
   \State
   \textbf{Return} $z_{t-1}$
  \end{algorithmic}
\end{algorithm}

\section{Experiments}
In this section, we provide extensive qualitative and quantitative comparisons to demonstrate the effectiveness of our approach, please refer to the appendix for more experimental results and analysis.
\subsection{Experimental Setups}
\textbf{Datasets.}\quad
\begin{figure*}[t]
    \centering
 \includegraphics[width=1.0\textwidth]{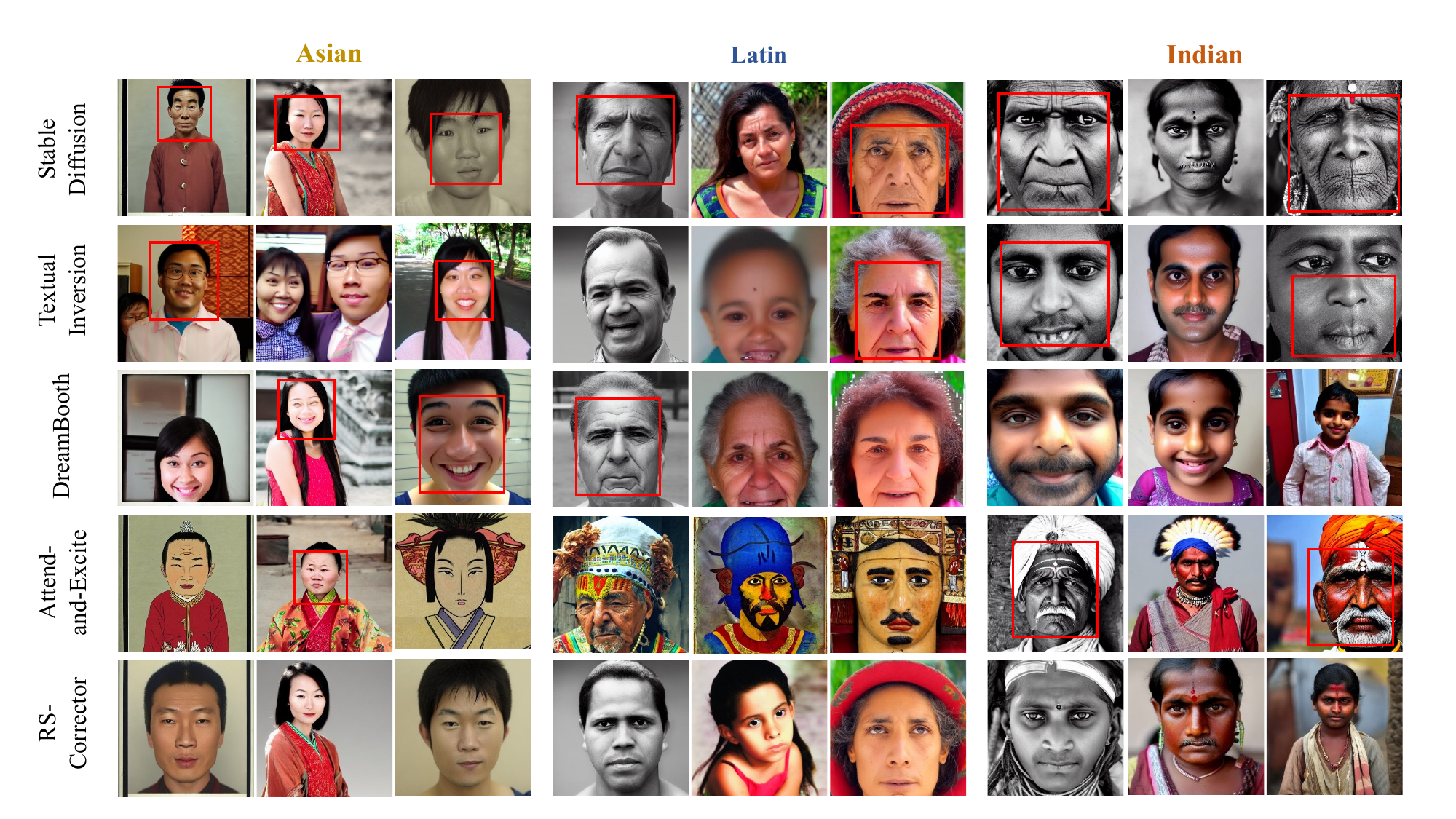}
    \caption{Qualitative comparison of generated results for the Asian, Latin, and Indian groups. For each racial group, we exhibit three images generated by comparing methods, where the generation seed is the same.}
    \label{fig:asian}
\end{figure*}
We establish a data platform to filter and re-annotate the facial images in LAION-5B~\cite{schuhmann2022laion}, considering racial groups of White, Black, Asian, Latin, and Indian.  Each racial group consists equally distributed gender and age groups to eliminate their influence. The conditioned textual prompts include ``a face of a white person'', ``a face of a black person'',  ``a face of an Asian'', ``a face of a Latin person'', and ``a face of an Indian''. We utilize a pre-trained face detection model libface\footnote{\url{ https://github.com/ShiqiYu/libfacedetection}} to identify and extract facial images from LAION-5B, which are subsequently labeled by CLIP~\cite{radford2021learning} roughly. Following the initial labeling, each image undergoes verification by human workers for more precise annotation. Images in the relabeled dataset are utilized in our approach as anti-stereotype samples.

\textbf{Implementation Details.}\quad
The projector $F(\cdot)$ is trained for 4 epochs with a batch size of 4 and a temperature parameter $\tau$ of 0.1. During inference, we experimentally choose the hyper-parameters with $\lambda_{tc}=9$, $\lambda_c=150$. The timesteps for updating the latent code range from $t_i=30$ to $t_k=26$ out of 30 total timesteps, with an optimization step size $\eta$ of 2. All the experiments are conducted on a single Tesla V100 GPU.

\textbf{Evaluation Metrics.} 
Traditional text-image similarity is conducted to evaluate the alignment between generated results and textual descriptions. However, considering the potential biases of CLIP inherited from the training data~\cite{radford2021learning,karakas2022fairstyle}, We employ Face++ API\footnote{\url{https://www.faceplusplus.com}} for more detailed evaluation. The evaluation metrics encompass the following two aspects: (i) Facial Aesthetics, including beauty and skin status assessments, is to evaluate whether specific racial groups are portrayed as degradation or unhealthy representation, considering the implications of racial stigmatization and vilification. 
(ii) Attribute Fairness, to evaluate whether the generated results exhibit potential biases across different racial groups, concerning attributes including age and gender. This metric is assessed by age consistency across different racial groups and gender distance of two types of gender. More detailed definitions of the adopted metrics can be referenced in the appendix.
\subsection{Qualitative Analysis}
\begin{table*}[h]
    \centering
    \resizebox{\linewidth}{!}{
    \begin{tabular}{c|rrrrr|rrrrr|rrrrr|r}
    % \toprule
    \Xhline{1pt}
    Metric & \multicolumn{5}{c|}{Beauty ($\uparrow$)}            & \multicolumn{5}{c|}{Skin Health ($\uparrow$)}      & \multicolumn{5}{c|}{Skin Defects ($\downarrow$)} &   
    % \multirow{2}*{\multicolumn{1}{c}{\makecell[c]{Time(s)\\($\downarrow$)}}}    \\ 
\multirow{2}{*}{\makecell[c]{Time(s)\\($\downarrow$)}} \\
    % \cmidrule{1-16}
    \Xcline{1-16}{0.4pt}
    Race   & \multicolumn{1}{c}{White}   & \multicolumn{1}{c}{Black} & \multicolumn{1}{c}{Asian}   & \multicolumn{1}{c}{Latin}   & \multicolumn{1}{c|}{Indian} & \multicolumn{1}{c}{White}   & \multicolumn{1}{c}{Black} & \multicolumn{1}{c}{Asian}   & \multicolumn{1}{c}{Latin}   & \multicolumn{1}{c|}{Indian} & \multicolumn{1}{c}{White}   & \multicolumn{1}{c}{Black} & \multicolumn{1}{c}{Asian}   & \multicolumn{1}{c}{Latin}   & \multicolumn{1}{c|}{Indian}  & \\ 
    % \midrule
    \Xhline{1pt}
    Stable Diffusion & 64.293  & 54.917  & 54.454  & 58.371  & 58.743  & 18.223  & 13.598  & 32.214  & 12.050  & 5.253   & 20.310  & 20.249  & 12.426  & 27.396  & 52.975 &  6.277  \\ 
    \Xhline{0.4pt}
    Textual Inversion & \underline{64.838}  & 49.030  & \underline{56.775}  & 60.289  & 57.535  & \underline{20.086}  & 6.439   & 27.186  & \underline{28.412}  & \textbf{15.055}  & 22.896  & 35.708  & \underline{11.037}  & \underline{17.649}  & \textbf{27.791}  &  \underline{9.190}  \\ 
    DreamBooth & 61.234  & 54.729  & 51.387  & 52.827  & \underline{59.881}  & 14.924  & 17.462  & 25.726  & 8.295   & 7.681   & \underline{21.903}  & \underline{13.487}  & 15.033  & 28.160  & 44.208 & 29.618   \\ 
    Attend-and-Excite & 64.390  & \textbf{63.210}  & 51.132  & \textbf{72.315}  & 56.713  & 16.315  & \underline{20.739}  & \underline{29.554}  & 11.807  & 8.406   & 27.475  & 24.673  & 17.770  & 39.332  & 50.445   & 15.700  \\ 
    RS-Corrector & \textbf{68.826}  & \underline{58.071}  & \textbf{62.044}  & \underline{63.928}  & \textbf{61.536}  & \textbf{49.805}  & \textbf{33.639}  & \textbf{53.852}  & \textbf{34.372}  & \underline{11.534}  & \textbf{7.290}   & \textbf{7.406}   & \textbf{6.808}   & \textbf{10.271}  & \underline{38.816}   &  \textbf{7.964}  \\ 
    % \bottomrule
    \Xhline{1pt}
    \end{tabular}}
    \caption{Facial Aesthetics evaluation and inference time comparison between RS-Corrector and others. The best performance is highlighted in \textbf{bold} and the second to best is highlighted by \underline{underlines}.}
    \label{table:face aesthetics}
\end{table*}
\begin{figure}[t]
    \centering
    \includegraphics[width=\linewidth]{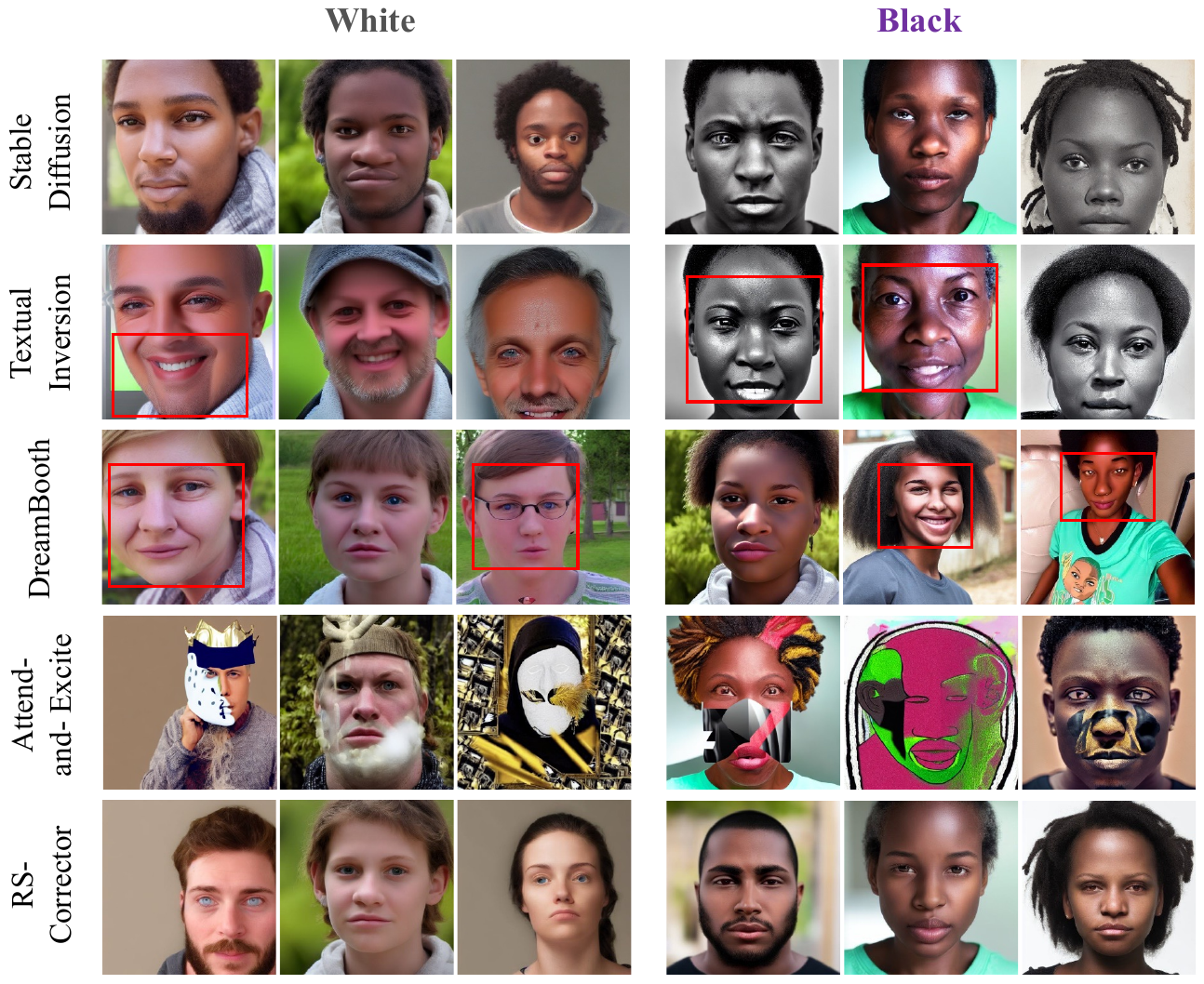}
    \caption{Additional comparisons for White and Black groups. We show three images generated by five methods, where the generation seed is the same.}
    \label{fig:black_white}
\end{figure}
\begin{figure}[t]
    \centering
    \includegraphics[width=\linewidth]{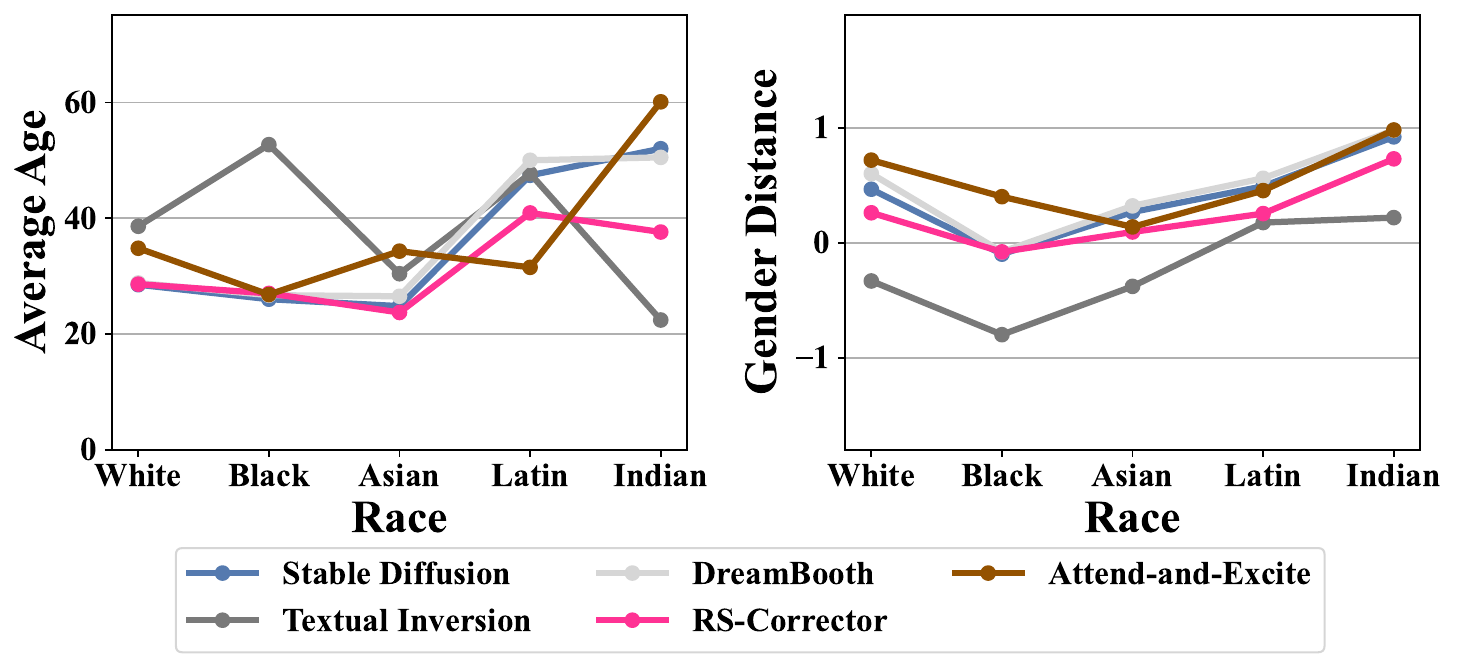}
    \caption{ Comparisons for average age and gender distance (calculated by subtracting the proportion of women from the proportion of men) consistency of different methods. More consistent values across different racial groups indicate better performance in attribute fairness. Our method showcases superior performance with respect to age and gender in comparison.}
    \label{fig:fig-age-gender}
\end{figure}

We perform the qualitative evaluations with the state-of-the-art latent diffusion models, Stable Diffusion~\cite{rombach2022high}, and its derivatives, including Textual Inversion~\cite{gal2022image}, DreamBooth~\cite{ruiz2023dreambooth} and Attend-and-Excited~\cite{chefer2023attend}. For a fair comparison, Textual Inversion and DreamBooth are fine-tuned on our dataset. Attend-and-Excite updates the latent code through the inference stage, with no need for fine-tuning.

In the first columns corresponding to the Asian group in Fig.~\ref{fig:asian},
Stable Diffusion tends to depict this group with disproportionately fixed small eyes, single eyelids, and excessively high cheekbones, which exhibit an unattractive appearance and raise concerns about potential racial prejudice and discrimination. 
Textual Inversion and DreamBooth, which fine-tune the original model, can generate images dissimilar to the SD results. However, their results suffer from discernible distortions, blurry content, and unnatural facial expressions (highlighted with red boxes). Since these models are designed for obtaining a new concept or learning a style from a few training samples, when confronted with diverse and attribute-rich training data, they are susceptible to disrupted final results (highlighted with red boxes). Moreover, the generated results of Attend-and-Excite tend to display stylistic features rather than authentic, also resembling the original SD. This resemblance is observed in fixed discriminatory facial characteristics reinforcing the persistence of racial stereotypes in the original model. In contrast, RS-corrector successfully rectifies the racial stereotype inherent in the original SD model, resulting in a significant improvement in aesthetic appearance. 
  
For the Latin and Indian racial groups, the generated results of SD appear to be excessively aged in comparison to other racial groups, exhibiting increased wrinkles, skin sagging, and skin discolorations, which indicates unfair performance in terms of age. Textual Inversion and Dreambooth demonstrate comparable aging patterns for the Latin group, deviating from other racial groups with respect to age. This inconsistency suggests that these models may not effectively identify the stereotypes present in the generated results and the generated results remain unimproved consequently. Attend-and-Excite also inherits the aging patterns for these racial groups. However, RS-corrector depicts a consistent age pattern of different groups, and the aesthetic appearance is also improved. Moreover, in Fig.~\ref{fig:black_white}, \themodel demonstrates superior improvements in aesthetic appearance compared to other methods the black and white groups, whereas . Overall, our method showcases its enhanced capability to mitigate stereotypes and improve the visual quality of the generated images across various racial groups compared to the state-of-the-art methods.

\subsection{Quantitative Analysis}
In Table~\ref{table:face aesthetics}, we present objective measurements of facial aesthetic. Compared to other methods, RS-Corrector achieves the best performance in most cases across all the racial groups. However, the results of Stable Diffusion for the black and Asian groups are less appealing compared to other racial groups. Textual Inversion and DreamBooth exhibit inconsistent improvements in face aesthetics, as their disrupted results may introduce additional defects, attenuating the overall performance. Attend-and-Excite can generate images with high beauty scores, as their results are stylistic and visually appealing for face aesthetics. However, it is hard for the model to yield authentic images and their skin status scores are also suboptimal. RS-Corrector achieves the best facial aesthetics performance in terms of beauty score and skin status. Specifically, the beauty score for the Asian group exhibits significant improvement, and the disparity between Asian and other racial groups is also reduced, demonstrating our refinement ability and efficacy in mitigating racial vilification. Note that the generated results of our method for the Indian group contain some religious ornaments, such as facial paints and feathers, which may have an impact on skin status scores for this racial group.

For attribute fairness evaluation, in Fig.~\ref{fig:fig-age-gender}, the average age and gender distance consistency is evaluated. Most models exhibit aging issues in the Indian and Latin groups compared to other racial groups. Additionally, Textual Inversion also manifests aging concerns in the black group. In contrast, our method exhibits the most consistent average age across different races, indicating superior age fairness. 
In terms of gender, Stable Diffusion tends to produce results with a higher proportion of men than women in most cases. Textual Inversion mitigates this gender bias for the Latin and Indian groups but remains unfair for other racial groups. DreamBooth and Attend-and-Excite exhibit similar patterns to Stable Diffusion. On the contrary, RS-Corrector demonstrates the best gender consistency performance across all the racial groups. Overall, our method showcases superior fairness in generating results for different racial groups concerning age and gender.

We also evaluated the efficiency of our method and others in terms of inference time, illustrated in Tab.~\ref{table:face aesthetics}. RS-Corrector exhibits superior efficiency compared to other models. Additionally, there is only a marginal increase of the inference time compared to the baseline Stable Diffusion model. This is because we only update the latent code during a limited number of timesteps and the overall inference timesteps are few as well. In contrast, other models necessitate more intricate computational processes and a greater number of inference timesteps, resulting in lower efficiency during inference.

\subsection{Ablation Study}
\begin{figure}[t]
    \centering
    \includegraphics[width=\linewidth]{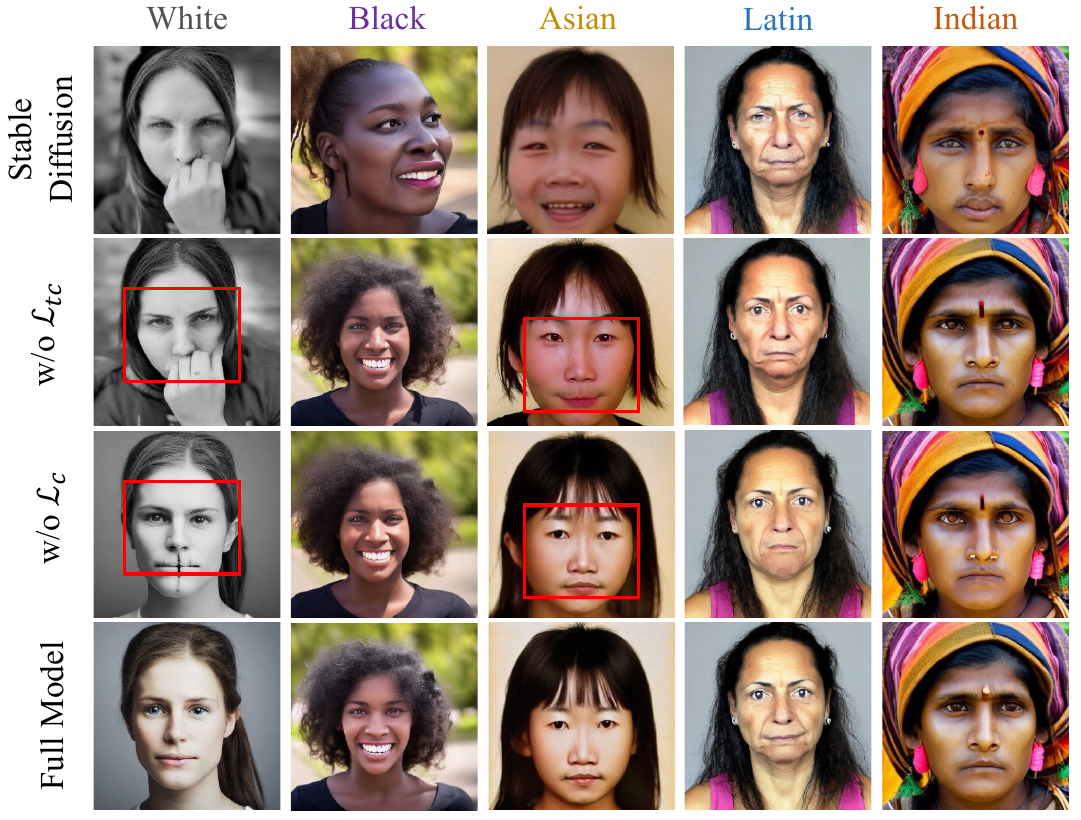}
    \caption{The results of the ablation study on RS-Corrector.}
    \label{fig:ab}
\end{figure}
\begin{figure}
    \centering
    \includegraphics[width=\linewidth]{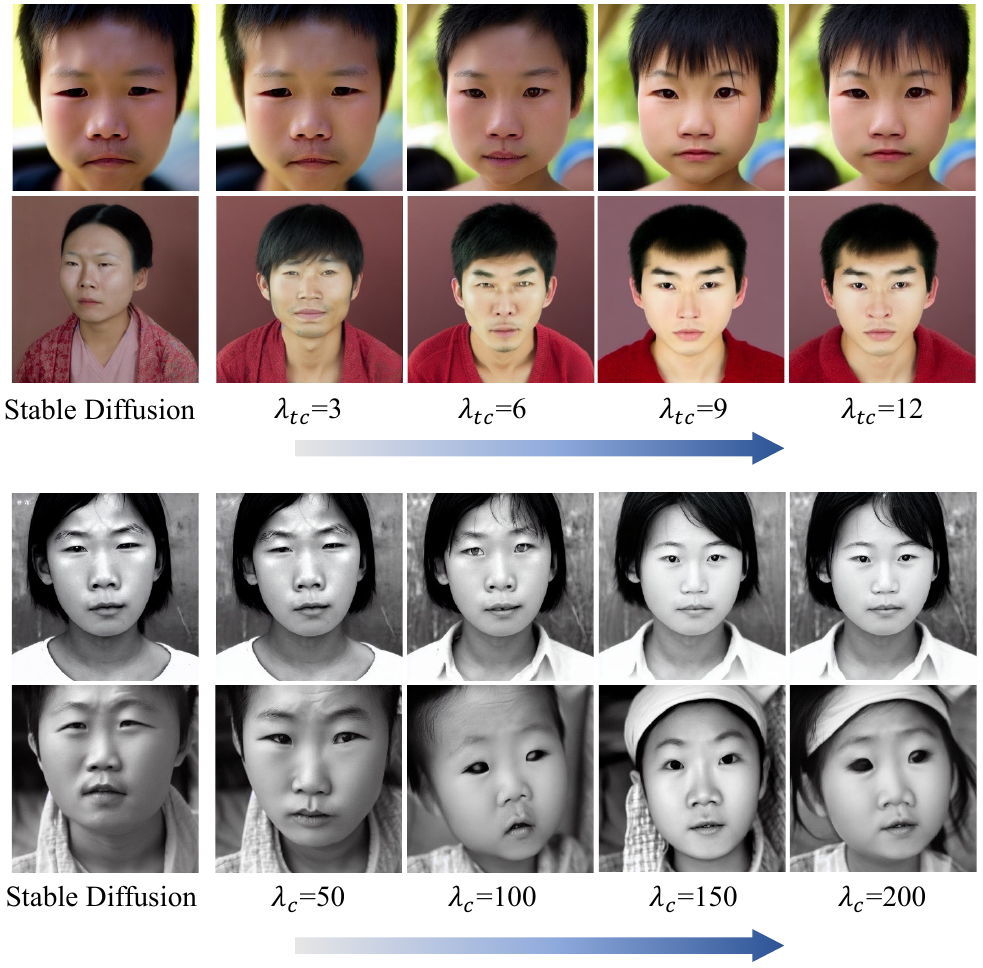}
    \caption{Visual comparisons of RS-Corrector with its variants.}
    \label{fig:losses}
\end{figure}

We show the ablation study of different latent constraints in Fig.~\ref{fig:ab}. The anti-stereotypical latent prior $\mathcal{L}_{tc}$ effectively guides the refinement in the overall structure of the original image, while direct anti-stereotypical guidance $\mathcal{L}_c$ refines facial features to enhanced aesthetic appeal while likely to maintain the structure of the original image. However, the full model demonstrate the best performance on facial appearance. To further understand how these constraints affect the updating process of the latent code, we separately conduct experiments for different configurations of each loss in Fig.~\ref{fig:losses}. 
% The first column shows the original images generated by SD, and the right columns are the results of different loss configurations. 
As the hyper-parameters increase, the refinement induced by the corresponding constraints on the final result becomes more pronounced and satisfying. However, when the increase exceeds a certain limit, the effectiveness begins to diminish.

\section{Conclusion}
In this paper, we introduce a simple yet effective framework called ``RS-Corrector'' to correct the racial stereotypes of latent diffusion models. Our approach aims to obtain an anti-stereotypical preference in that latent space and update the latent code for refined generated results. We establish a latent prior to distinguish the preference, cooperating with the direct anti-stereotypical guidance in a contrastive manner, which is obtained during the inference stage from anti-stereotypical and stereotypical samples. Extensive experiments demonstrate our method achieves superior refinement and fair results compared with the state-of-the-art methods.

\clearpage
{
    \small
    \bibliographystyle{ieeenat_fullname}
    \bibliography{main}
}


\clearpage
\setcounter{page}{1}
% \maketitlesupplementary
\appendix
% In addition to the experiments and discussions presented in the main paper, the supplementary furnishes more implementation details and experimental results, which further demonstrate the effectiveness of our method.

% \section{Rationale}
% \section*{Appendix}D
\section{Implementation Details}
\subsection{Additional Explanation}
This subsection further explains the evaluation metrics referenced in the paper.

\textbf{Facial Aesthetics.}\quad The facial authenticity assessment is undertaken through the Face++ API\footnote{\url{https://www.faceplusplus.com}}, as delineated in the study. Concretely, we adopt three metrics for facial authenticity assessment. (i) Beauty Score, referenced in the paper, signifies the mean value of male-assessed beauty score and female-assessed beauty score. (ii) Skin Health, represents the health condition of the images related to skin characteristics. (iii) Skin Defects, including stains, acne, and dark circles. The score referenced in the paper signifies the mean value of the above assessments. Overall, higher beauty and skin health scores, and lower skin defects scores denote a more appealing and attractive facial morphology of generated results.

\textbf{Gender Distance.}\quad Given a gender signal $g_i$ ($i$ is the number of generated facial images), $g_{i}=1$ indicates the gender is male and $g_{i}=-1$ indicates female. The Gender Distance score is calculated as follow:
\begin{equation}
    \begin{aligned}
        % score=\frac{\sum Male - \sum Female}{\sum Male + \sum Female}.
        % score=\frac{\sum_i g_i }{\sum_i |g_i|}.
        score=\frac{\sum_i g_i }{\sum_i |g_i|}.
    \end{aligned}
\end{equation}
Lower scores indicate a more equitable distribution between males and females in generated images, while heightened score stability signifies greater gender consistency across different racial groups.
\section{Additional Quantitative Results}\quad
\subsection{Additional Metrics}
To enhance the assessment of the efficacy of our method, we further extend our evaluation to additional quantitative metrics.
% As a supplement to the paper, we quantitatively evaluate both race accuracy and generation diversity.

\textbf{Race Accuracy.}\quad
Race accuracy (Race Acc), is determined by the correct classification of all generated results according to specific racial groups. Given the potential disparity between the generated results and the specified racial attributes, we conduct an evaluation to verify the accuracy of the generated races, which is predicted through race analysis using Deepface\footnote{\url{https://github.com/serengil/deepface}}.

  \begin{figure}[t]
    \centering
 \includegraphics[width=1.0\linewidth]{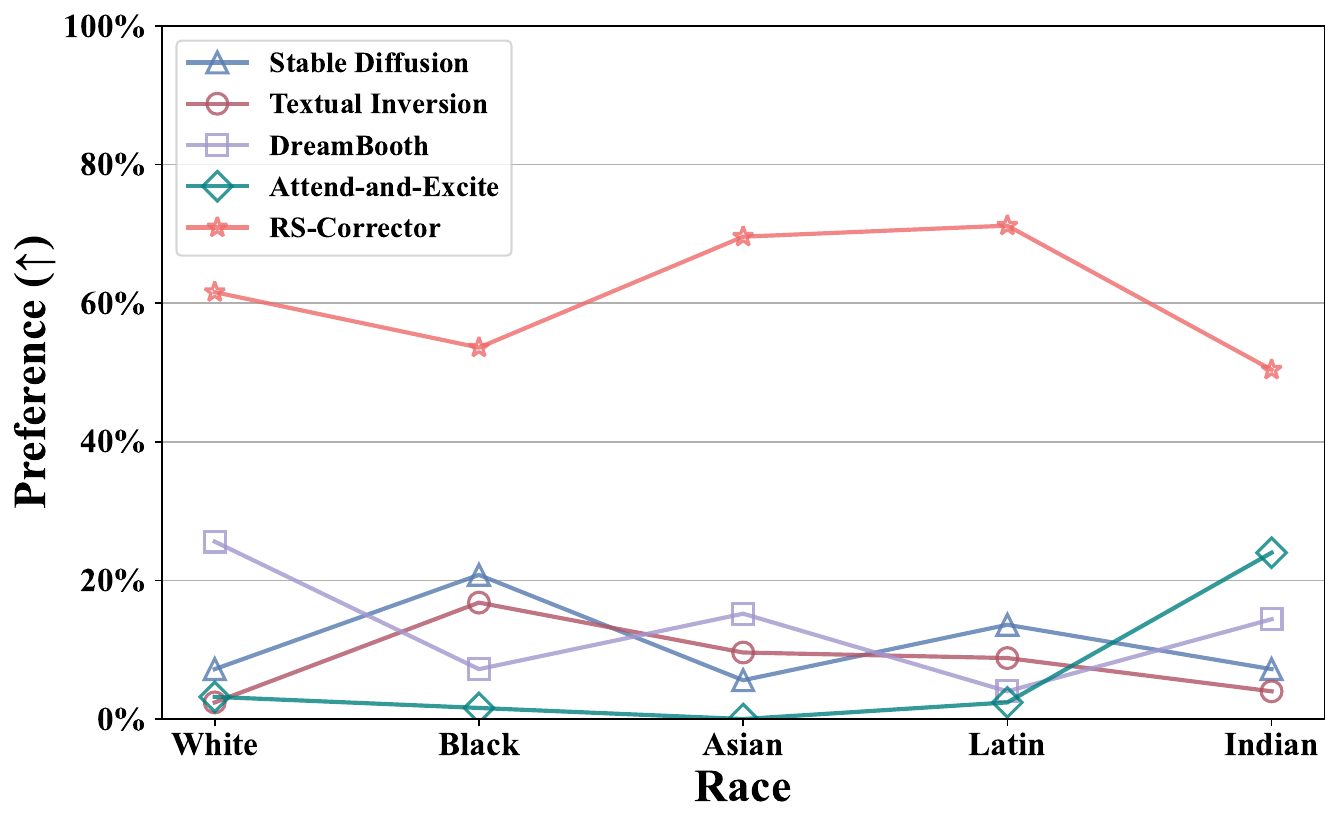}
    \caption{User study. The results show the average percentage of cases in which the results of different methods are preferred for each racial group.}
    \label{fig:ref}
\end{figure}

\begin{table}[t]
\centering
    \resizebox{\linewidth}{!}{
\begin{tabular}{c|rrrrr}
    \toprule
Metric & \multicolumn{5}{c}{Race Acc ($\uparrow$)}        \\ \hline
Race & White & Black & Asian & Latin & Indian \\ \hline
        Stable Diffusion    & 0.586 & 0.839 & 0.891 & \underline{0.437} & \textbf{0.864} \\
        Textual Inversion   & 0.722 & \underline{0.852} & 0.796 & 0.370 & \underline{0.863} \\
        DreamBooth           & \underline{0.856} & 0.753 & 0.707 & 0.148 & 0.468 \\
        Attend-and-Excite   & 0.755 & 0.709 & \underline{0.909} & 0.173 & 0.815 \\
        RS-Corrector         & \textbf{0.874} & \textbf{0.926} & \textbf{0.948} & \textbf{0.586} & 0.854 \\
\bottomrule
\end{tabular}
}
\caption{Race accuracy evaluation for different methods. The best performance is highlighted in \textbf{bold} and the second to best is highlighted by \underline{underlines}.}
    \label{table:race_acc}
\end{table}

\begin{table}[t]
\centering
    \resizebox{\linewidth}{!}{
\begin{tabular}{c|rrrrr}
    \toprule
Metric & \multicolumn{5}{c}{Intra-LPIPS ($\uparrow$)}        \\ \hline
Race & White & Black & Asian & Latin & Indian \\ \hline
        Stable Diffusion & 0.575 & 0.543 & 0.644 & 0.662 & 0.664 \\
        Textual Inversion & 0.613 & 0.629 & 0.649 & 0.594 & 0.651 \\
        DreamBooth & 0.614 & 0.600 & 0.639 & 0.636 & 0.658 \\
        Attend-and-Excite & \textbf{0.737} & \textbf{0.730} & \textbf{0.710} & \textbf{0.786} & \textbf{0.717} \\
        RS-Corrector & \underline{0.630} & \underline{0.655} & \underline{0.664} & \underline{0.692} & \underline{0.680} \\
\bottomrule
\end{tabular}
}
\caption{ Quantitative results of intra-LPIPS for different methods. Higher intra-LPIPS indicates better performance on generation diversity. Attend-and-Excite demonstrates commendable performance, however, it falls short in addressing the essential challenge of generating diversity in authentic racial images. For authentic racial images, our method achieves the best performance.}
    \label{table:gen_div}
\end{table}

\begin{table*}[t]
\centering
    % \tiny
    % \resizebox{\textwidth}{!}{
\begin{tabular}{c|ccccc}
    \toprule
Method & Stable Diffusion & Textual Inversion & DreamBooth &  Attend-and-Excite & RS-Corrector \\ \hline
Preference ($\uparrow$) & 10.88\% & 8.32
\% & 13.28\% & 6.24\% & \textbf{61.28\%}  \\
\bottomrule
\end{tabular}
% }
\caption{User study. The results show the average preference across five racial groups of different methods. The best results are in \textbf{bold}.}
    \label{table:us}
\end{table*}

\textbf{Intra-cluster LPIPS.}\quad
Intra-cluster LPIPS (Intra-LPIPS), indicates the generation diversity of compared methods. Fine-tuning on a specific dataset may lead to overfitting, causing fine-tune-based methods to struggle in generating diverse images. To assess the diversity of the generated results and investigate potential overfitting in compared methods, we employ intra-LPIPS following~\cite{ojha2021few,zhao2022closer}. $K$ cluster-center images are first selected and other generated images are assigned to one of the centers. The intra-LPIPS is obtained by computing
the average standard LPIPS~\cite{zhang2018unreasonable} for random paired images within
each cluster, followed by averaging this value over $K$ clusters.

\begin{table}[t]
\centering
    % \resizebox{\linewidth}{!}{
\begin{tabular}{r|rrr}
    \toprule
 & w/o $\mathcal{L}_{tc}$ & w/o $\mathcal{L}_{c}$ & Full Model  \\ \hline
        Beauty ($\uparrow$) &55.920&59.369&\textbf{62.044}  \\
        Skin Health ($\uparrow$)&37.242&45.481&\textbf{53.852} \\
        Skin Defects ($\downarrow$)&10.821&8.499&\textbf{6.808} \\
        Race Acc ($\uparrow$) & 0.895 & 0.900 &\textbf{0.930} \\
        Intra-LPIPS($\uparrow$) & 0.640 & 0.649 & \textbf{0.661} \\
\bottomrule
\end{tabular}
% }
\caption{Ablation study on different loss terms.}
    \label{table:ab}
\end{table}

\subsection{Quantitative Results}
In Table~\ref{table:race_acc}, we present the quantitative measurement of race accuracy for comparison. The results are averaged over 500 generated images for each racial group. Compared with the state-of-the-art methods, we perform the best race accuracy in most cases. Additionally, the results for generation diversity are presented in Tab.~\ref{table:gen_div}, similarly averaging over 500 generated images for each method. As the results demonstrate, Attend-and-Excite exhibits enhanced performance in generation diversity. This is attributed to its inclination to generate stylized images for most racial groups (in Fig.~\ref{fig:W}, Fig.~\ref{fig:B}, Fig.~\ref{fig:y}, Fig.~\ref{fig:L}), thereby contributing to increased generation diversity.  However, it encounters difficulties in producing authentic racial images, which essentially challenges generative models on racial stereotypes. In addressing this pivotal facet while preserving generation diversity, our approach consistently achieves optimal performance.
% \subsection{qualitative results}
\section{User Study}
We conducted a survey with 25 participants to assess the preferences for generated images of each method. Participants were tasked to select images perceived to exhibit the least racial prejudice, discrimination, and vilification. A total of 625 votes were collected from the participants, representing assessments across five racial groups. The preferences across five racial groups for each method are illustrated in Fig.\ref{fig:ref}, and the average preferences of different racial groups are presented in Tab.\ref{table:us}. In both results, our method outperforms other state-of-the-art methods, demonstrating our superior performance at generating anti-stereotypical racial images by human evaluation.

\section{Quantitative Results of Ablation Study}
The quantitative results of the ablation study are present in Tab.~\ref{table:ab} for metrics referenced in the paper and supplementary. As the results show, omitting either $\mathcal L_{tc}$ or $\mathcal L_c$ leads to a decline in performance across facial aesthetics, race accuracy, and generation diversity. In contrast, the full model demonstrates superior performance across all quantitative metrics.

\section{Additional Qualitative Results}
\begin{figure*}[t]
    \centering
 \includegraphics[width=1.0\textwidth]{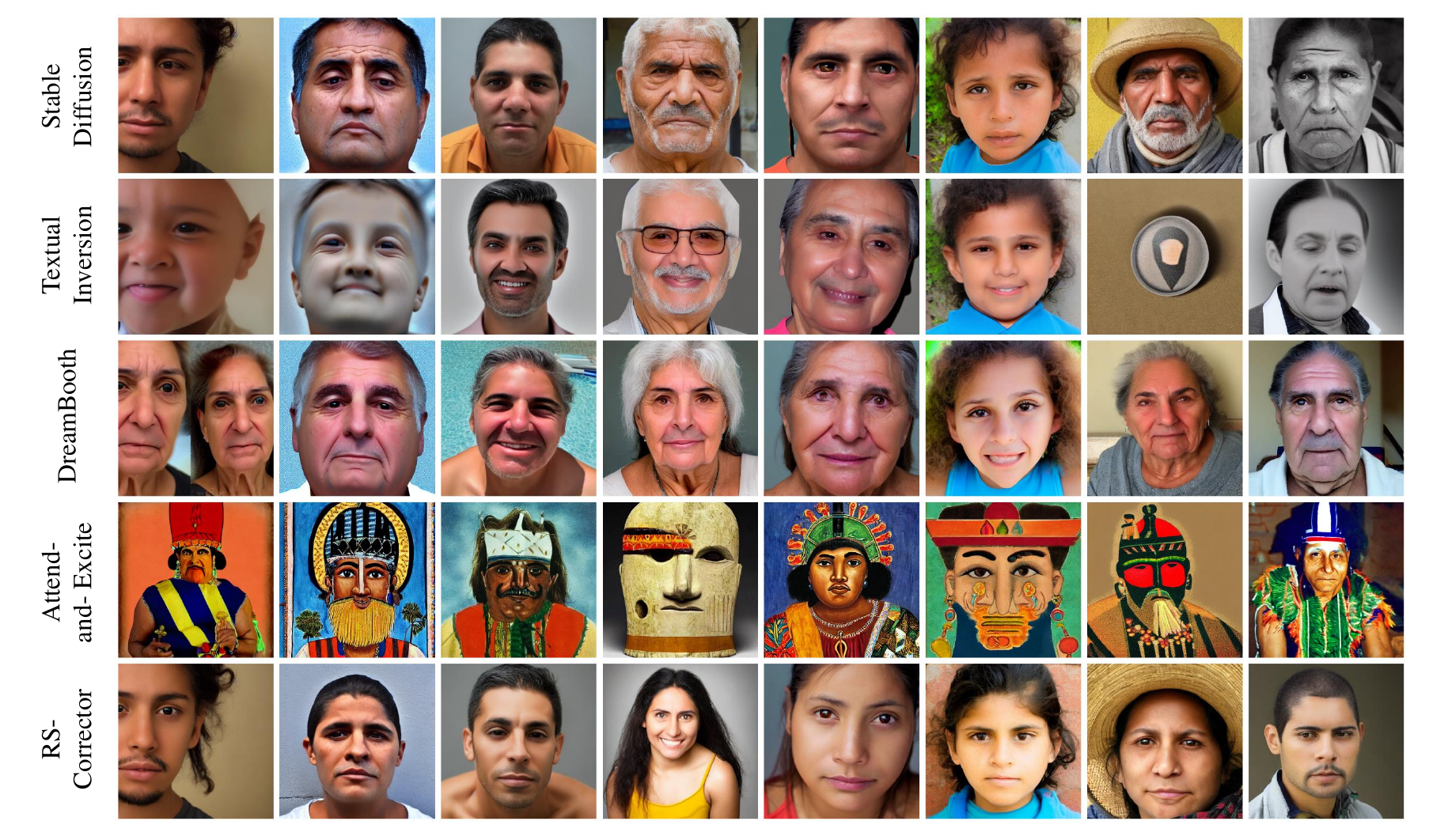}
    \caption{Qualitative comparison of generated results for the Latin racial group. The generation seed is the same for different methods.}
    \label{fig:L}
\end{figure*}

\begin{figure*}[t]
    \centering
 \includegraphics[width=1.0\textwidth]{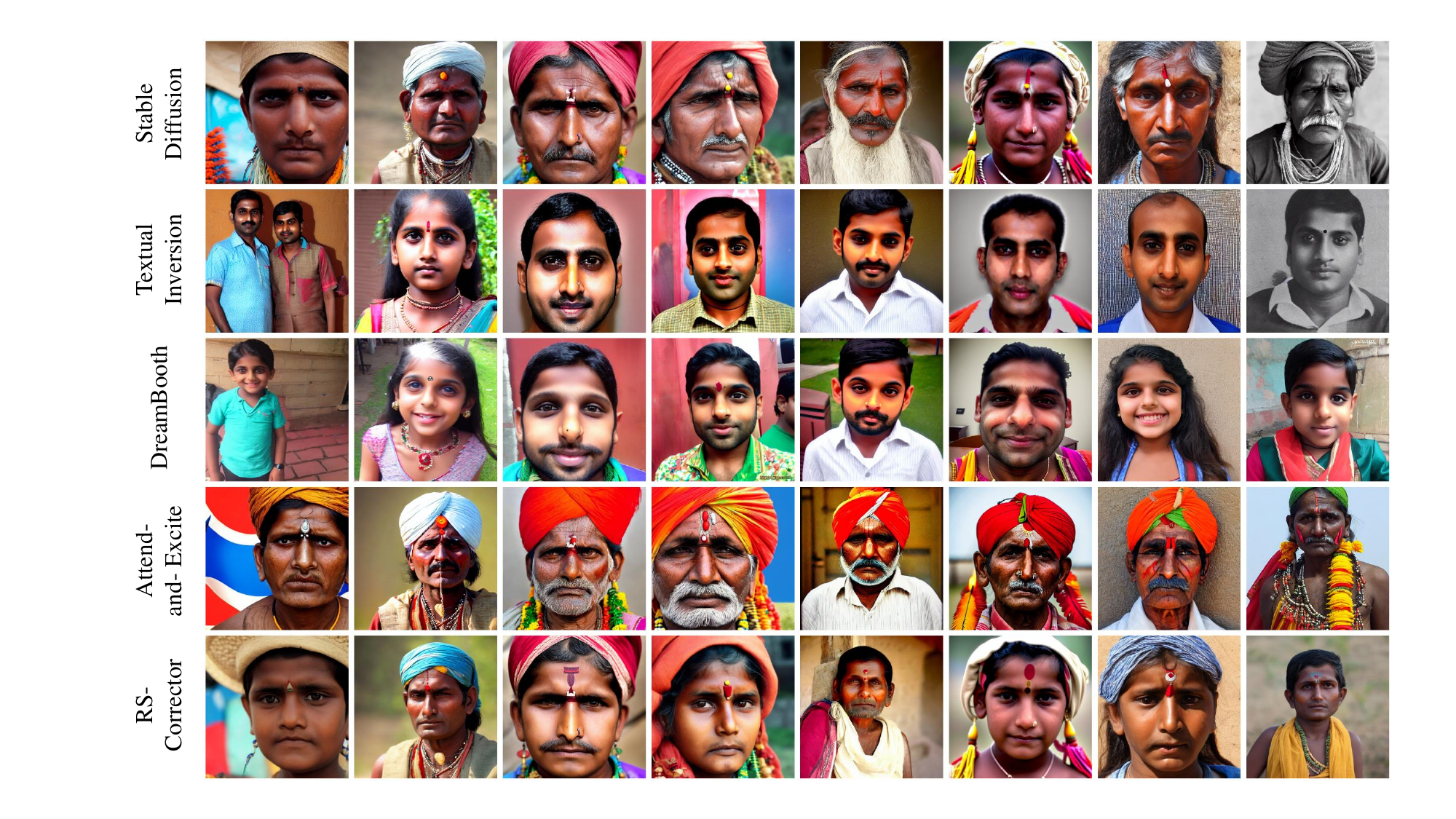}
    \caption{Qualitative comparison of generated results for the Indian racial group. The generation seed is the same for different methods.}
    \label{fig:I}
\end{figure*}

 As a supplement to the paper, we present additional qualitative results of our method on five racial groups compared with other methods in the following figures (Fig.~\ref{fig:W}, Fig.~\ref{fig:B}, Fig.~\ref{fig:y}, Fig.~\ref{fig:L}, Fig.~\ref{fig:I}). The experimental results further demonstrate the effectiveness of our method and the out-performance compared to the SOTA works.
 
 \begin{figure*}[t]
    \centering
 \includegraphics[width=1.0\textwidth]{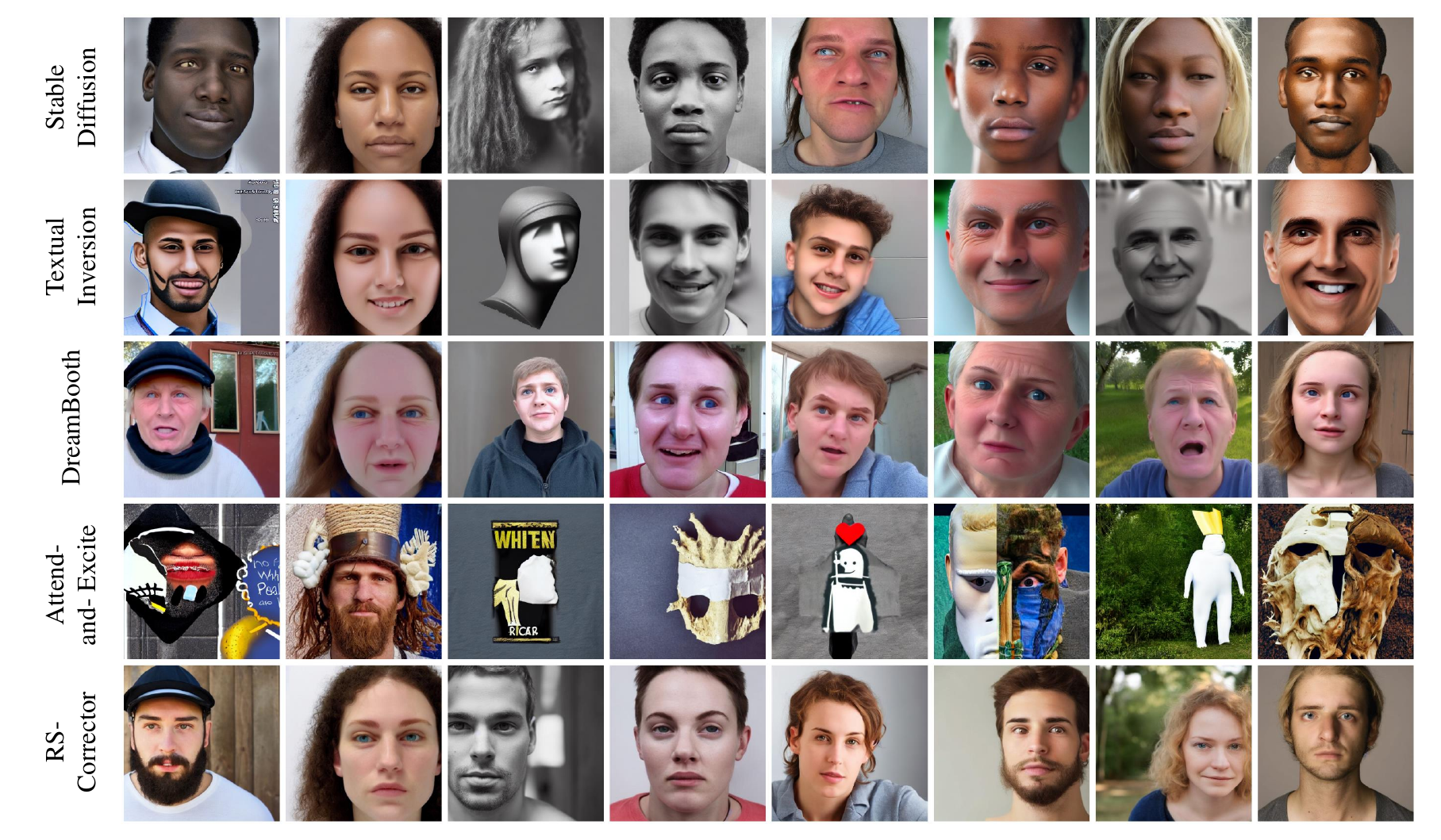}
    \caption{Qualitative comparison of generated results for the White racial group. The generation seed is the same for different methods.}
    \label{fig:W}
\end{figure*}

\begin{figure*}[t]
    \centering
 \includegraphics[width=1.0\textwidth]{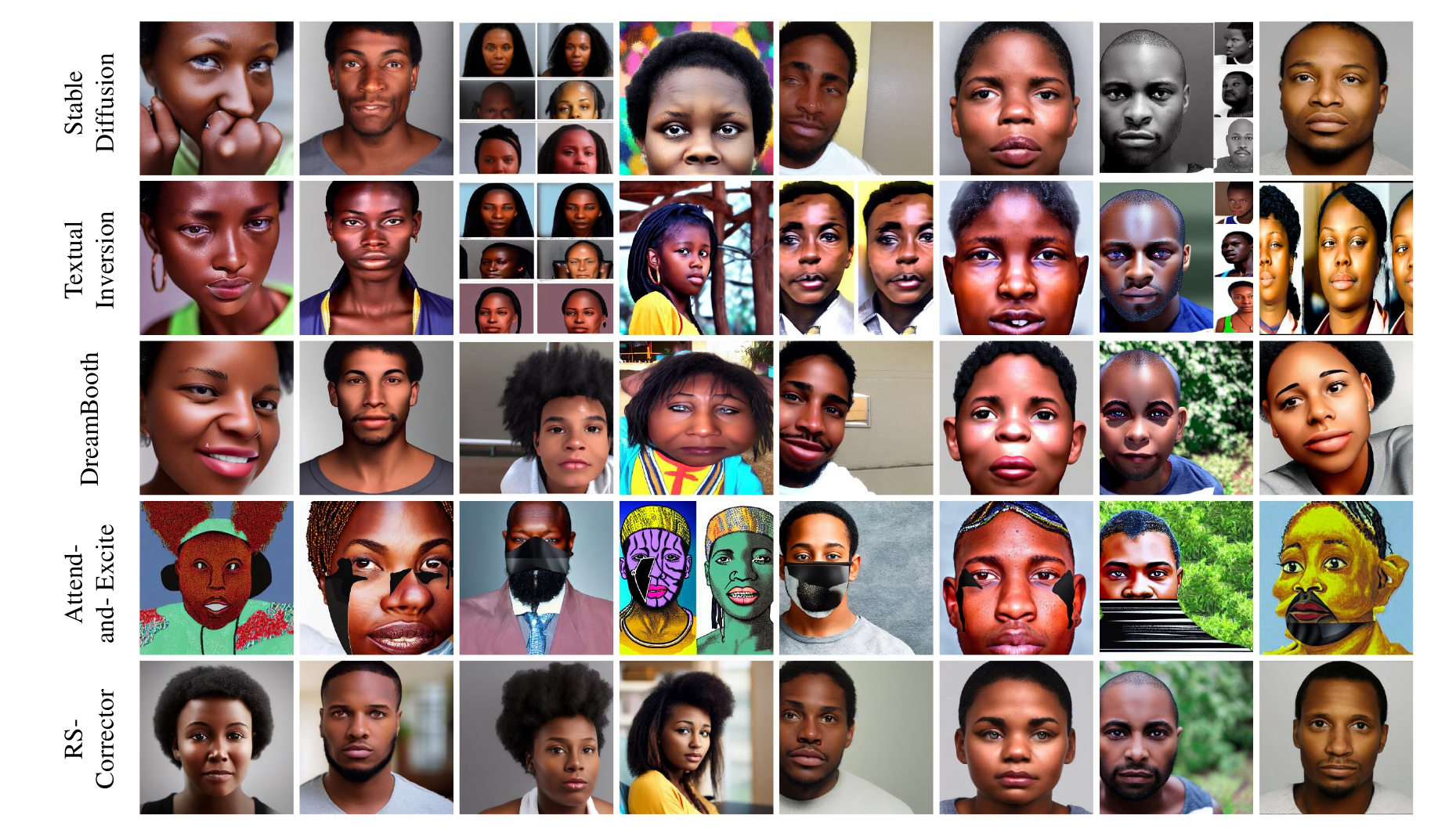}
    \caption{Qualitative comparison of generated results for the Black racial group. The generation seed is the same for different methods.}
    \label{fig:B}
\end{figure*}

\begin{figure*}[t]
    \centering
 \includegraphics[width=1.0\textwidth]{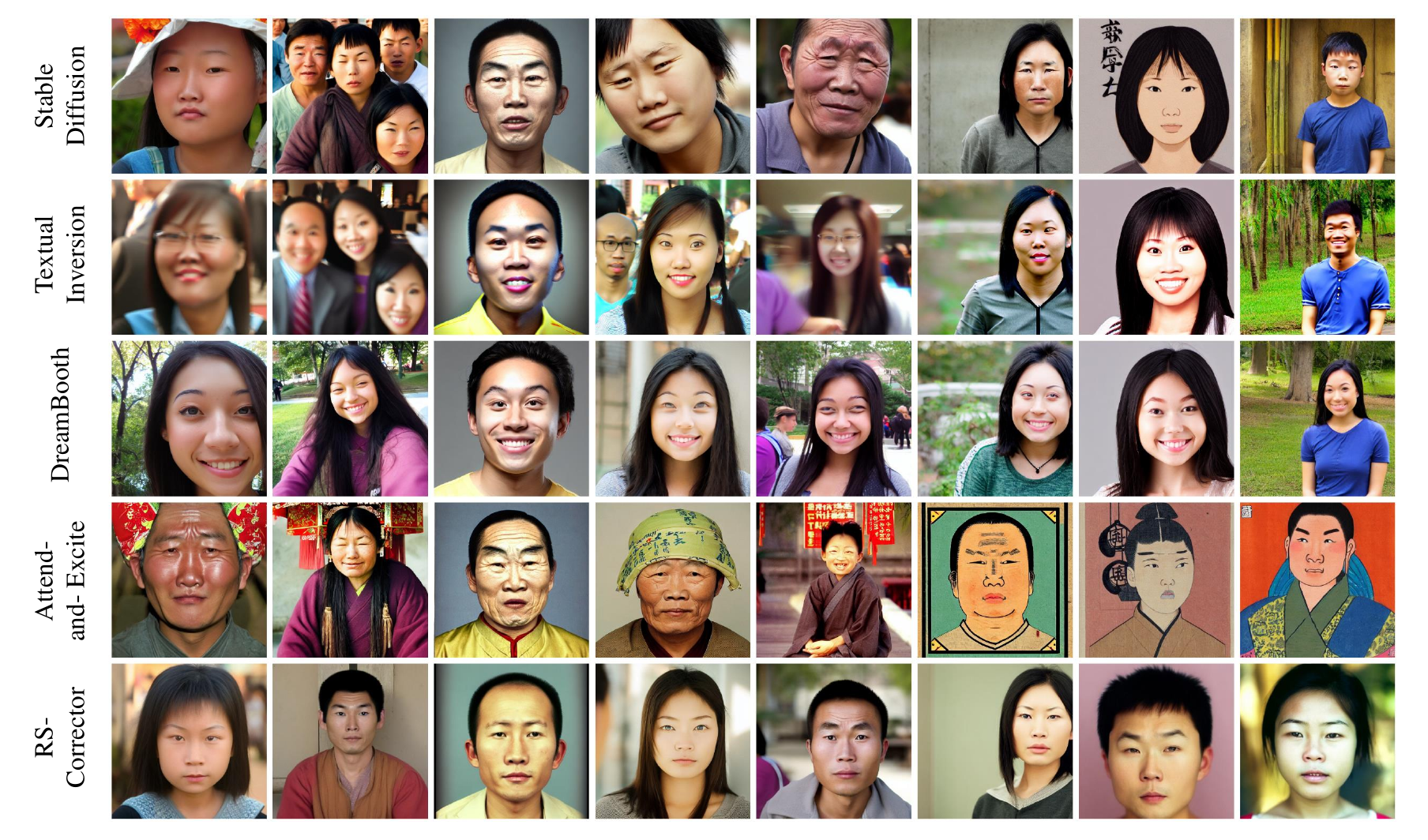}
    \caption{Qualitative comparison of generated results for the Asian racial group. The generation seed is the same for different methods.}
    \label{fig:y}
\end{figure*}

% \label{sec:rationale}
% 看一下格式
% 
% 

\end{document}